\documentclass[runningheads]{llncs}
\usepackage[T1]{fontenc}

\usepackage{graphicx}
\usepackage{amsmath}
\usepackage{amssymb} 
\usepackage{booktabs}
\usepackage[margin=1in]{geometry}
\usepackage{array}
\usepackage[linesnumbered,ruled,vlined]{algorithm2e}
\usepackage{fontawesome}
\usepackage{xcolor}
\usepackage{subcaption}
\usepackage{placeins}  
\usepackage{float}  
\begin{document}
\title{A Deep-Learning Framework for Land-Sliding Classification from Remote Sensing Image}

\author{Quang-Hieu Tang\inst{1}\orcidID{0009-0006-7922-4040} \and
Nhat-Truong Vo Dinh\inst{2}\orcidID{0009-0002-8190-891X} \and
Dong-Dong Pham\inst{3}\orcidID{0009-0003-1236-3925} \and 
Quoc-Toan Nguyen\inst{4}\orcidID{0000-0001-9256-7087} \and
Lam Pham\inst{5}\orcidID{0000-0001-8155-7553} \and
Truong Nguyen\inst{6}\orcidID{0000-0002-7422-4600}
}
\authorrunning{Hieu et al.}
%
\institute{University of technology of Troyes, Troyes 10300, France \\
\email{hieu.tang@utt.fr}\and
ICC Dong Duong, Ho Chi Minh City, Vietnam\\
\email{truongvnd277@gmail.com}\and
Biosel AI, Ho Chi Minh City, Vietnam\\
\email{dongpd.work@gmail.com}\and
University of Technology Sydney, Australia\\
\email{quoctoan.nguyen@student.uts.edu.au}\and
Austrian Institute of Technology\\
\email{lam.pham@ait.ac.at} \and
Ho Chi Minh City University of Technology (HCMUT)\\
\email{truongnguyen@hcmut.edu.vn}}
\maketitle              
\begin{abstract}
The use of satellite imagery combined with deep learning to support automatic landslide detection is becoming increasingly widespread. However, selecting an appropriate deep learning architecture to optimize performance while avoiding overfitting remains a critical challenge. To address these issues, we propose a deep-learning based framework for landslide detection from remote sensing image in this paper. 
The proposed framework presents an effective combination of the online an offline data augmentation to tackle the imbalanced data, a backbone EfficientNet\_Large deep learning model for extracting robust embedding features, and a post-processing SVM classifier to balance and enhance the classification performance. The proposed model achieved an F1-score of 0.8938 on the public test set of the Zindi challenge. 

\end{abstract}

\section{Introduction}
Landslides are increasingly becoming a widespread natural disaster, driven by various objective factors such as global warming and human-induced geological surface modifications. 
Such incidents have brought growing attention to the study of landslides, particularly in regions with a history of geo-hazards.
Historically, landslide analysis has primarily relied on two conventional approaches: (i) geo morphological field surveys and (ii) visual interpretation of aerial imagery or orthophotographs \cite{moha21}. 
The method (i) involves in-situ field mapping, often conducted post disaster such as following seismic events to validate and complement remote sensing data, including satellite imagery. 
Meanwhile, the method (ii) typically utilizes stereo pairs of aerial photographs, some dating back to the 1950s, to assess changes in vegetation cover, terrain features, and both seasonal and long-term landscape dynamics. 
Both of two conventional methods are inherently time-consuming and require substantial domain expertise. As a result, their application in real-time disaster prediction is limited, often hindering rapid response and risking the omission of affected regions.

With the advancement of satellite technologies, multi-spectral satellite imagery now offers spatial resolutions as fine as 10 meter per pixel~\cite{nasa} and provides a wide range of spectral bands. 
These bands provides detailed characterization of surface properties such as moisture, vegetation cover, and topographical structure. 
Leveraging this information, scientists can generate comprehensive landslide risk maps and compile extensive geo-spatial datasets documenting geological structural changes over decades.
To analyze multi-spectral satellite imagery, machine learning and deep learning techniques have demonstrated significant success. Classical machine learning algorithms such as Random Forest, Rotation Forest, Support Vector Machine (SVM), and Logistic Regression have been extensively employed for generating landslide susceptibility maps~\cite{kavz19}. 
Recently, deep-learning based methods have been increasingly utilized for landslide classification tasks. For example, Liu et al.\cite{liu21} conducted a comparative study of three convolutional neural network architectures including CNN, ResNet, and DenseNet for landslide classification. Lam et al.\cite{rmaunet} and 
Soares et al.\cite{soar20} explored the effects of patch size and sampling strategy on the performance of a UNet model applied to landslide segmentation. Long et al.\cite{long21} proposed a hybrid framework that integrates a Deep Belief Network (DBN) and a CNN in parallel, fusing their extracted features prior to classification in order to enhance remote sensing-based landslide detection performance. 
Furthermore, Qin et al.\cite{qin21} introduced a novel domain transfer learning approach by incorporating a maximum mean discrepancy (MMD) loss function to bridge the distribution gap between source and target domains.

Despite the deep-learning models present advances, several research gaps still remain. In particular, existing models implicitly assume that landslide samples are sufficiently represented in the dataset. In practice, however, landslide occurrences are rare compared to non-landslide regions, resulting in highly imbalanced data distributions. Due to the imbalanced data and the limitation of landslide region, deep learning models with excessive complexity often exhibit limited generalizability. 
In this paper, we propose an end-to-end deep-learning framework for landslide classification which aim to address the class imbalance problem.
To this end,  in the pre-processing phase, we employ multiple data augmentation such as the Synthetic Minority Oversampling Technique (SMOTE~\cite{smote}), cut mix,  mixup, etc. to generate synthetic minority samples to enrich the training dataset. 
Then, we construct a deep neural network architecture, which leverages a pre-trained EfficientNetV2-Large model, for the training process. Finally, in the post-processing stage, we replace the classification head of the neural network by a Support Vector Machine (SVM) layer to enhance classification fairness and overall effectiveness.

\section{Background}

\subsubsection{Synthetic Minority Oversampling Technique (SMOTE):}
In supervised learning tasks with imbalanced datasets, detection models tend to be biased towards the majority class (i.e., nonlandslide regions).
This leads poor detection performance for the minority class (i.e., landslide regions). 
The imbalance is particularly problematic in classification tasks.
To mitigate this issue, the Synthetic Minority Oversampling Technique (SMOTE), which is widely used for oversampling, generates synthetic training samples for the minority class. Unlike naive oversampling methods that duplicate existing instances, SMOTE synthesizes new samples by interpolating between a sample and its \(k\)-nearest neighbors (typically \(k=5\)) within the minority class in the feature space.
Formally, given a minority class instance \(\textbf{X}_i\) and one of its neighbors \(\textbf{X}_{i}^{(nn)}\), a synthetic instance is generated as: $\textbf{X}_{\text{s}} = \textbf{X}_i + \lambda \cdot (\textbf{X}_{i}^{(nn)} - \textbf{X}_i)$ where \(\lambda \in [0, 1]\) is a random scalar.
The SMOTE algorithm is outlined as follows:

\begin{algorithm}[H]
\caption{SMOTE with Structural Similarity for Multi-Band Images}
\KwIn{
    Minority class image set $\mathcal{I} = \{I_1, I_2, \dots, I_N\}$ with multiple bands\\
    Number of neighbors $k = 5$\\
    Number of synthetic samples per image: $n_{\text{syn}}$\\
    Clipping bounds for interpolation weight: $[a, b] = [0.1, 0.9]$\\
    Beta distribution parameters: $\alpha$, $\beta$
}
\KwOut{Augmented minority class set $\mathcal{I}'$ with synthetic images}

\ForEach{image $I \in \mathcal{I}$}{
    Compute SSIM between $I$ and all other images in $\mathcal{I}$\;
    Identify $k = 5$ nearest neighbors $\mathcal{N}_I = \{I_1', I_2', \dots, I_5'\}$ based on SSIM distance\;
    
    \For{$j = 1$ \KwTo $n_{\text{syn}}$}{
        Randomly select $I_n \in \mathcal{N}_I$\;
        Sample interpolation weight $\lambda \sim \text{Beta}(\alpha, \beta)$, and clip: $\lambda \leftarrow \text{clip}(\lambda, a, b)$\;
        Generate synthetic image: $\tilde{I} = \lambda I + (1 - \lambda) I_n$\;
        Add $\tilde{I}$ to augmented set $\mathcal{I}'$\;
    }
}
Return $\mathcal{I}'$
\end{algorithm}

\subsubsection{Cut Mix:} This method operates by replacing a random rectangular region of an image with a corresponding patch from another image. The label of the resulting mixed image is then computed as a weighted combination of the original labels, proportional to the area of the respective image contributions. Let the dataset consist of input-label pairs $\mathcal{X} = \{(X_0, L_0), \dots, (X_{n-1}, L_{n-1})\}$ where $X_i \in \mathbb{R}^{W \times H \times C}$, $L_i$ is the label. The CutMix procedure can be formally described as follows: 








\begin{algorithm}[H]
\caption{CutMix Data Augmentation}
\label{alg:cutmix}
\KwIn{Dataset $\mathcal{X} = \{(X_0, L_0), \dots, (X_{n-1}, L_{n-1})\}$}
\KwOut{Augmented sample $(\tilde{X}, \tilde{L})$}

Randomly select two samples $(X_a, L_a)$ and $(X_b, L_b)$ from $\mathcal{X}$\;

Randomly generate rectangle coordinates $(x_1, y_1), (x_2, y_2)$ such that: \\
\Indp
$0 \leq x_1 < x_2 \leq W$, \quad $0 \leq y_1 < y_2 \leq H$\;
\Indm

Copy $X_a$ to $\tilde{X}$\;

Replace the region $[x_1:x_2, y_1:y_2]$ in $\tilde{X}$ with the same region from $X_b$\;

Compute mixing ratio: \\
\Indp
$\lambda = 1 - \dfrac{(x_2 - x_1)(y_2 - y_1)}{W \times H}$\;
\Indm

Compute the mixed label: \\
\Indp
$\tilde{L} = \lambda L_a + (1 - \lambda) L_b$\;
\Indm

\Return $(\tilde{X}, \tilde{L})$\;

\end{algorithm}

\subsubsection{Mixup:}

The Mixup technique is conceptually similar to the previously described CutMix method but differs in two key aspects. First, Mixup blends two entire images rather than localized regions. Second, the labels are combined using the same mixing ratio applied to the image data. The formulation of the Mixup algorithm is given in Equation~\eqref{eq:mixup}:

\begin{align}
    X_{\mathrm{Mixup}} &= \lambda X_a + (1 - \lambda) X_b \notag \\
    L_{\mathrm{Mixup}} &= \lambda L_a + (1 - \lambda) L_b
    \label{eq:mixup}
\end{align}

Here, $(X_a, L_a)$ and $(X_b, L_b)$ represent two randomly sampled image-label pairs from the dataset, and $\lambda \in [0, 1]$ is the mixing coefficient that controls the degree of interpolation between the two samples.
\subsubsection{EfficientNetV2-Large:}

\begin{figure}[ht]

    \centering
    \includegraphics[width=\linewidth]{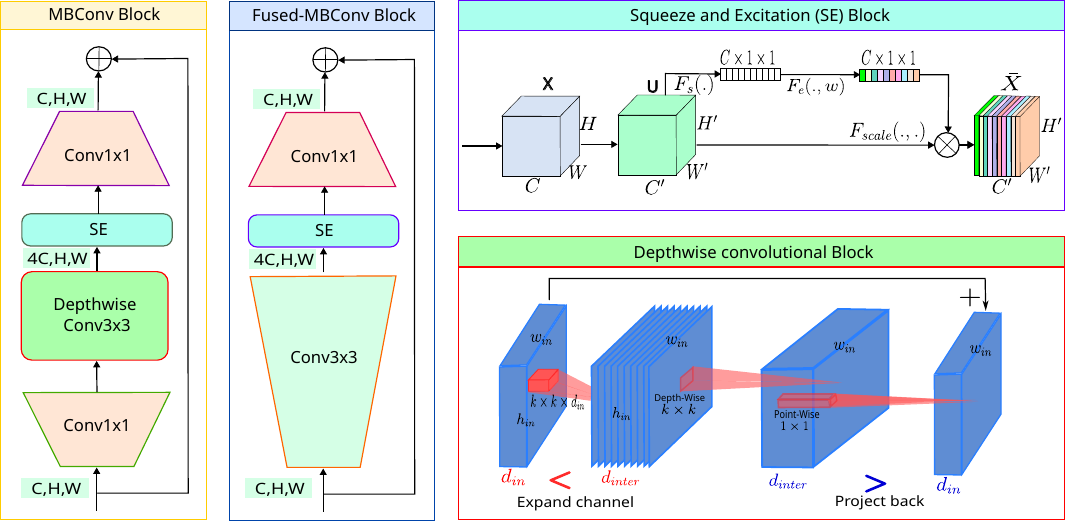}
    \caption{Unit blocks in the EfficientNetV2-Large network architecture}
    \label{fig:efficient_net_v2_large}
\end{figure}

EfficientNetV2-Large is belongs to the EfficientNetV2 family proposed by Tan and Le~\cite{effi21} (2021). 
This architecture is designed to address the trade-off between accuracy, parameter efficiency, and training/inference speed.
Three factors are crucial to deploy deep learning models in real-world scenarios with limited computational resources.
Therefore, we leverage the EfficientNetV2-Large network architecture for our proposed deep-learning framework for landslide classification in this paper.
Regarding EfficientNetV2 general network architecture, it presents a hybrid design combining both \textit{MBConv} and \textit{Fused-MBConv} building blocks. 
The MBConv block (Mobile Inverted Bottleneck Convolution~\cite{mobi18}) is a lightweight structure that includes depth-wise separable convolution and squeeze-and-excitation modules to enhance parameter efficiency. 
However, due to hardware inefficiencies in handling depthwise convolutions in the early layers, EfficientNetV2 introduces Fused-MBConv that replaces expansion and depthwise layers with a single standard 3×3 convolution followed by batch normalization and nonlinearity. 
This modification significantly improves training throughput, especially on accelerators such as GPUs or TPUs.
The architecture of EfficientNetV2-Large, as shown in Table~\ref{tab:efficientNetV2_Large_architecture} and Fig.~\ref{fig:efficient_net_v2_large}, follows the compound scaling principle in different way.
Unlike EfficientNetV1 models which applies this scaling parameter uniformly in all stages, EfficientNetV2-Large applies training-aware NAS framework~\cite{mnas2019} to seek optimal for convolutional operation types \{\textit{MBConv}, \textit{Fused-MBConv}\}, the number of layers, kernel size \{3x3, 5x5\}, and expansion ratio \{1, 4, 6\}. 
This is effective to mitigate the aggressively scale up image dimension in previous variants.

\begin{table}[h]
\centering
\caption{Layer settings of MBConv and Fused-MBConv blocks in EfficientNetV2-Large architecture}
\begin{tabular}{>{\raggedright\arraybackslash}p{1cm}  
    >{\raggedright\arraybackslash}p{4.5cm}  
    >{\centering\arraybackslash}p{1.5cm}  
    >{\centering\arraybackslash}p{1.5cm}  
    >{\centering\arraybackslash}p{2cm}  
    >{\centering\arraybackslash}p{2cm}  
    }
\toprule
\textbf{Stage} & \textbf{Operator} & \textbf{Expand} & \textbf{Stride} & \textbf{\#Channels} & \textbf{\#Layers} \\
\midrule
0 & Conv, k3×3                       & \_ & 2 &  32   & 1  \\
1 & Fused-MBConv, k3×3              & 1 & 1 &  32   & 4  \\
2 & Fused-MBConv, k3×3              & 4 & 2 &  64   & 7  \\
3 & Fused-MBConv, k3×3              & 4 & 2 &  96   & 7  \\
4 & MBConv, k3×3, SE               & 4 & 2 &  192  & 10  \\
5 & MBConv, k3×3, SE               & 6 & 1 &  224  & 19  \\
6 & MBConv, k3×3, SE               & 6 & 2 &  384  & 25 \\
7 & MBConv, k3×3, SE               & 6 & 1 &  640  & 7 \\
8 & Conv, k1x1 \& AvgPool \& FC      & \_ & \_ &  1280 & 1  \\
\bottomrule
\end{tabular}
\label{tab:efficientNetV2_Large_architecture}
\end{table}



\subsubsection{Support Vector Machine (SVM):}

Support Vector Machine (SVM) is a powerful and widely used supervised learning algorithm, especially effective in binary classification problems. 
The main objective of SVM is to find the optimal decision boundary, called a hyperplane, which separates data points of different classes with the maximum possible margin between them.

Assume that we have a binary classification dataset:
\[
\{(x_1, y_1), (x_2, y_2), ..., (x_n, y_n)\}, \quad x_i \in \mathbb{R}^d, \quad y_i \in \{-1, +1\}
\]
The goal of SVM is to find a hyperplane defined by a weight vector \(\mathbf{w}\) and bias \(b\) such that:
\begin{equation}
    \mathbf{w}^\top x + b = 0
    \label{eq:svm_equation}
\end{equation}
This hyperplane should separate the two classes such that all positive samples lie on one side and all negative samples lie on the other, with the largest possible margin. The margin is defined as the distance between the hyperplane and the closest points of each class (support vectors). Mathematically, we solve the following optimization problem:



\begin{equation}
\min_{\mathbf{w}, b, \xi} \ \frac{1}{2} \|\mathbf{w}\|^2 + C \sum_{i=1}^{n} \xi_i
\quad \text{subject to} \quad y_i(\mathbf{w}^\top x_i + b) \geq 1 - \xi_i, \quad \xi_i \geq 0\\ 
\label{eq:svm_regularization}
\end{equation}
where $C$ is the hyper parameter to control regularization strength.


\section{Our proposed deep-learning framework}
\label{proposed_system}

\begin{figure}
    \centering
    \includegraphics[width=\linewidth]{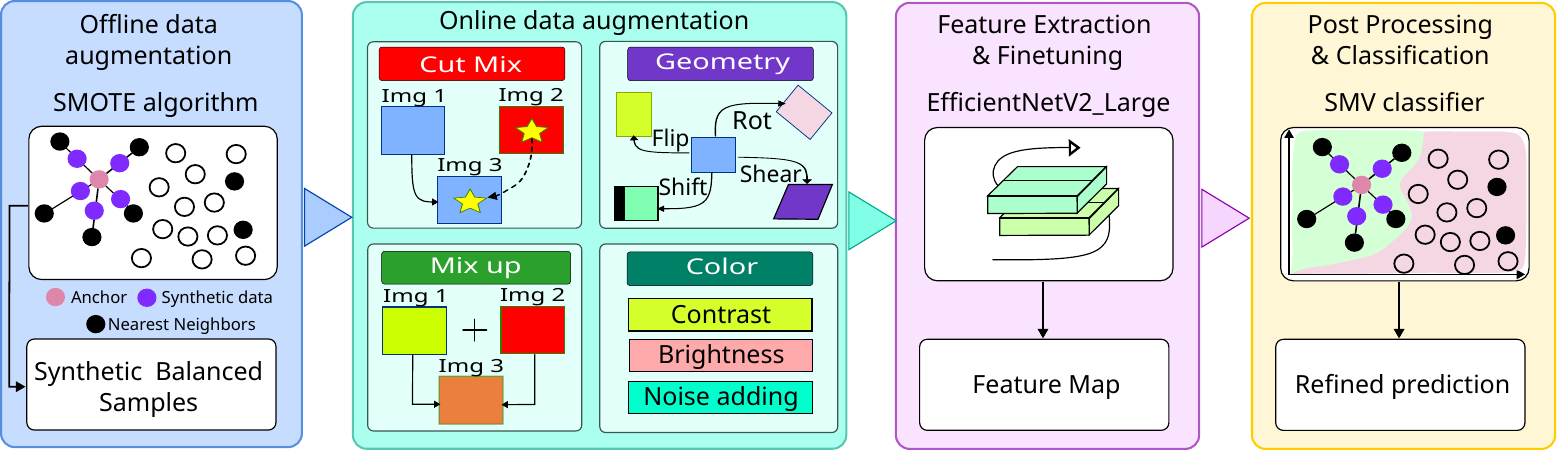}
    \caption{The high-level architecture of proposed deep-learning framework}
    \label{fig:overal_system}
\end{figure}

As shown in Fig.~\ref{fig:overal_system}, the high-level archtiecture of the proposed framework comprises three main parts: (1) Offline and online data augmentation, (2) Finetuning pre-trained deep neural network EfficientNetV2-Large, and (3) Post-processing on the feature map using SVM algorithm

\textbf{Data Augmentation:} First, SMOTE algorithm is used to generate more remote sensing images to tackle the imbalance data issue in this paper.
To this end, the structural similarity is employed as the metric to identify the five nearest neighbors for each minority-class image, referred to as the anchor image. For each anchor–neighbor pair, a blending weight is sampled from a Beta distribution and used to interpolate between the two images. This process generates synthetic samples that preserve structural coherence within the minority class. As SMOTE is applied to landslide image only and before the training process, we referred to as the off-line data augmentation.

Given the entire data after applying SMOTE, batches of data (i.e., a batch of data includes both landslide and none-landslide images) are performed to feed into the deep-learning model.
For each batch, we apply multiple data augmentations, referred to as the online data augmentation.
In particular, we adopt the \textbf{bag of freebies} strategy~\cite{bag_of_freebies}, which comprises three categories of transformations. 
First, \textbf{color transformations} are applied to each image in the training batch, including random noise injection, color jittering, and adjustments to saturation and contrast. Second, \textbf{geometric transformations} are employed, such as shifting, shearing, flipping, and rotation. 
Third, the \textbf{CutMix} and \textbf{Mixup} techniques, as previously described in this paper, are utilized to further enhance sample diversity without affecting inference cost. 


\textbf{Finetuning pre-trained EfficientNetV2\_Large network architecture:} In this paper, we leverage the EfficientNetV2\_Large architecture to extract deep embedding features from satellite imagery, referred to as the backbone EfficientNetV2\_Large.
Notably, the backbone EfficientNetV2\_Large was trained on the ImageNet dataset in advance.
Then, a linear fully connected (FC) layer is employed to predict the category of the input data.
The entire network is trained end-to-end using the backpropagation algorithm. 
In other words, all trainable parameters of the entire network is updated during the training process.

\begin{figure}[h]
    \centering
    \includegraphics[width=\linewidth]{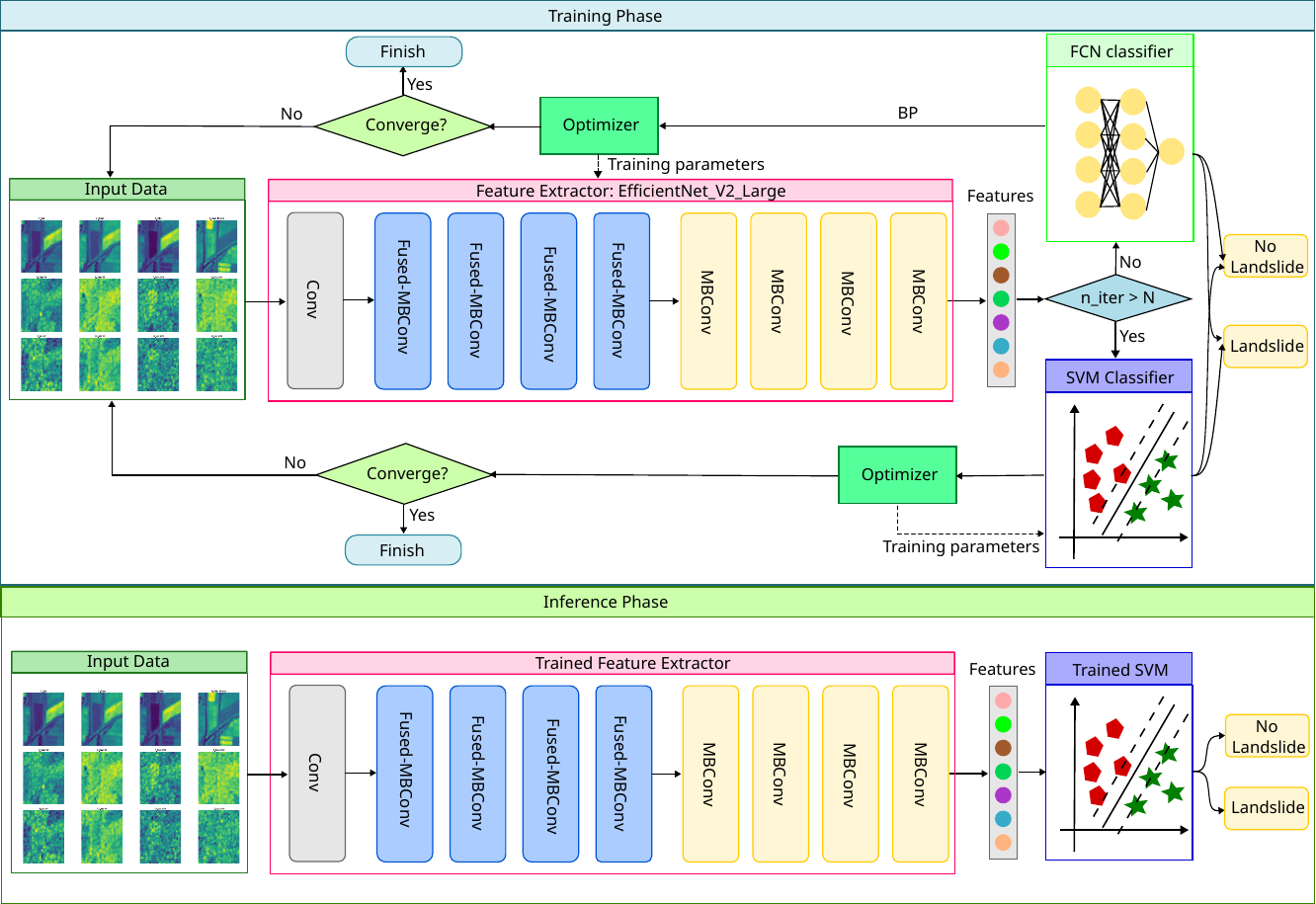}
    \caption{The training and inference processes of the proposed deep-learning framework}
    \label{fig:training_flow}
\end{figure}

\textbf{Post-processing using SVM:} In the training process, the model tends to overfit and becomes biased toward the majority class—namely, the Non-landslide class. 
To address this issue and enhance the classifier's fairness, we replace the FC head with a non-linear Support Vector Machine (SVM) classifier. This SVM, utilizing a Radial Basis Function (RBF) kernel and a regularization parameter $C$. This SVM serves two main purposes in our architecture: (i) As a lightweight baseline classifier to evaluate the separability of features learned by the backbone model; (ii) As a regularized decision boundary that is robust to outliers and small datasets, particularly important in landslide detection where positive samples are limited. 
By applying SVM for post-processing step, this is effective to evaluate the quality of learned embeddings while minimizing overfitting and reducing computational cost during inference. 
This approach ultimately leads to superior performance on the unseen test dataset.
As a result, the training and inference processes are illustrated in Fig. \ref{fig:training_flow}.

\begin{figure}[htbp]
    \centering
    \begin{subfigure}[b]{0.45\linewidth}
        \centering
        \includegraphics[width=\linewidth]{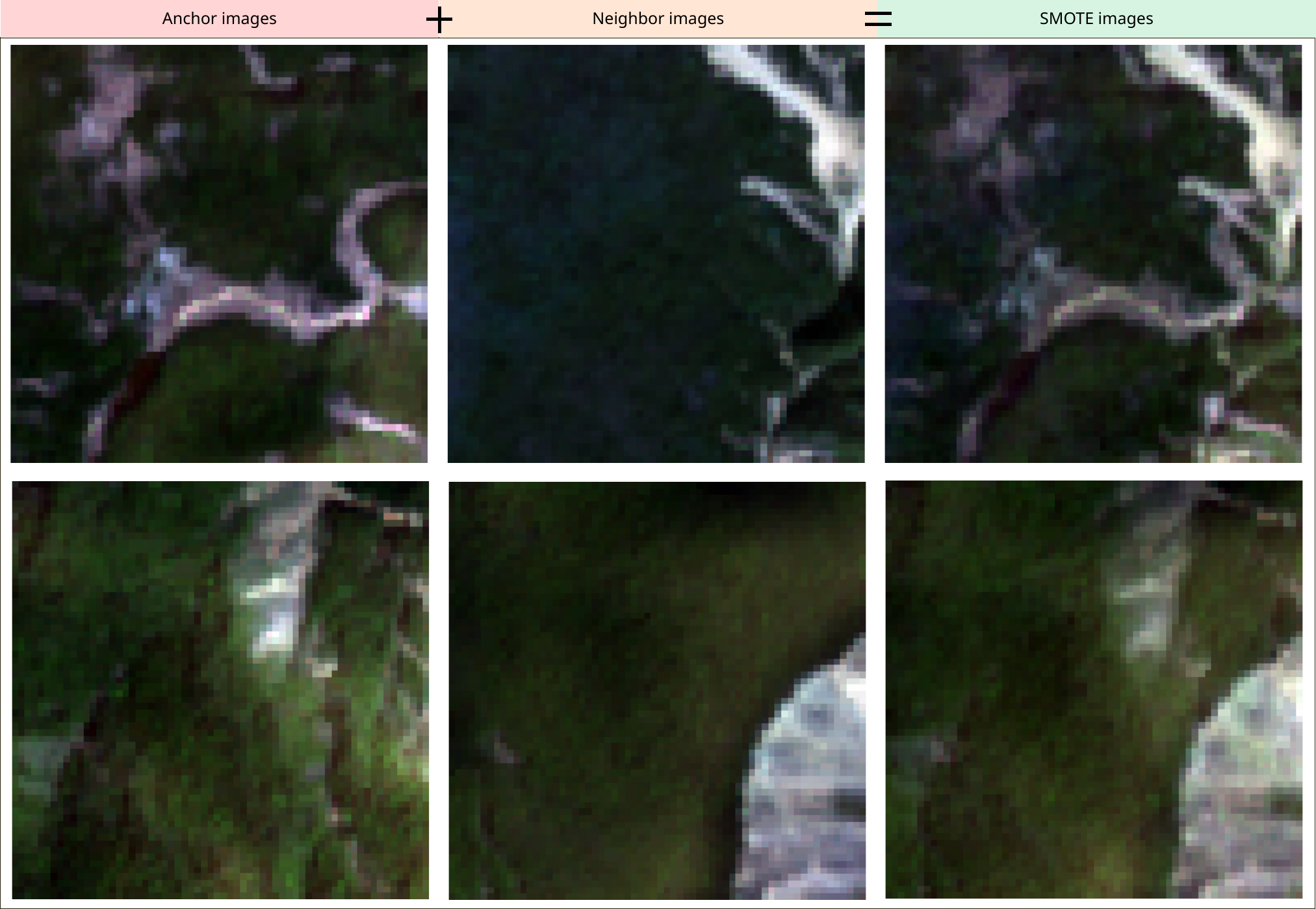}
        \caption{}
    \end{subfigure}
    \hfill
    \begin{subfigure}[b]{0.45\linewidth}
        \centering
        \includegraphics[width=\linewidth]{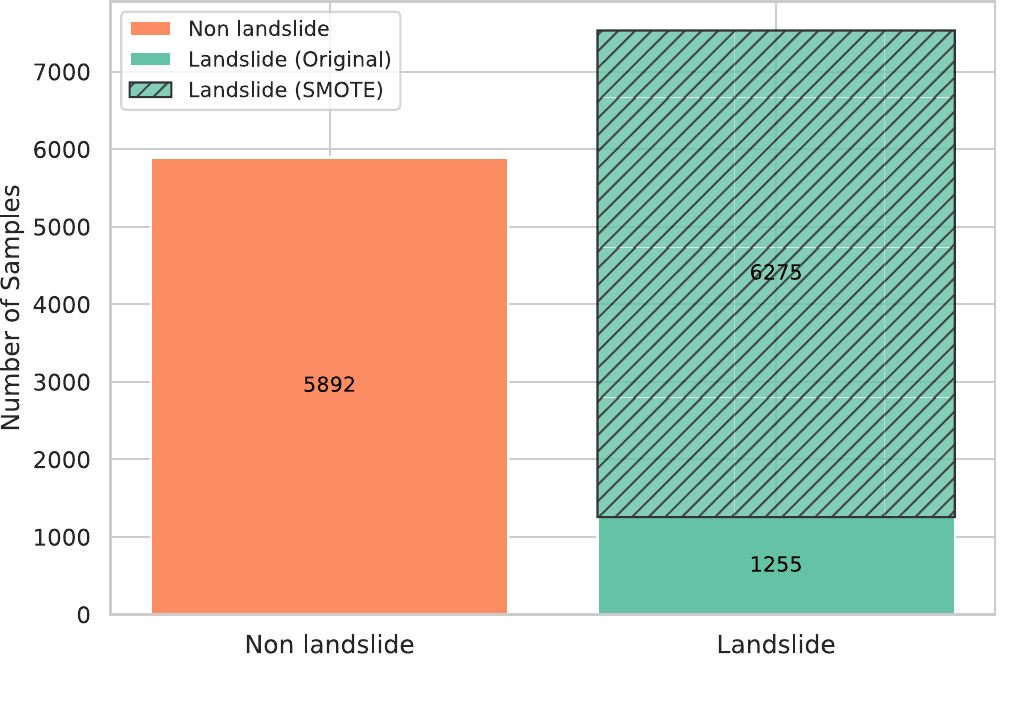}
        \caption{}
    \end{subfigure}
    \caption{(a) Left: Anchor images; Middle: Nearest neighbors; Right: SMOTE-generated samples. (b): Class distribution on the training subset after applying the SMOTE algorithm.}
    \label{fig:smote_images}
\end{figure}

\section{Experiments and Discussion}

\subsection{Zindi Challenge Dataset}
In this paper, we evaluate our proposed deep-learning framework on Zindi Challenge dataset.
This data was released as part of a data science competition hosted on the Zindi Africa platform~\cite{zindi}.
The challenge targets the problem of landslide detection using satellite-based remote sensing. 
It comprises multi-source Earth observation data, integrating both optical and radar imagery from the Sentinel-2 and Sentinel-1 satellite missions, respectively.
%
As mentioned in Section~\ref{proposed_system}, SMOTE algorithm is applied to Zindi Challenge dataset to generate more data.
As Fig.~\ref{fig:smote_images} shows, we generate more 6275 landslide images, enhancing the balance between landslide and none-landslide classes.

Each image in the dataset contains 28 channels, comprising optical bands (RGB, NIR), radar features (VV, VH, and derived change metrics), and handcrafted indices such as NDVI, EVI2, VARI, REDR, and various edge detectors (e.g., Sobel, Canny).
We adopt a 5-fold cross-validation strategy to ensure robustness and reduce bias in performance estimation. The training and testing image paths, along with feature statistics, are loaded from structured CSV and HDF5 files. 
The input images are resized to \(256 \times 256\), with standard scaling applied across features.
We follow the challenge rule. Then, we train and evaluate the proposed networks on the training set. Finally, we run the inference process on the test set, submit the results to the Leaderboard to obtain the test scores.
Notably, the challenge is still on going at the paper writing time.


\subsection{Experimental Settings}

Evaluating deep neural networks in this paper were constructed using PyTorch framework. 
All experiments were conducted on a single machine equipped with an  GPU GTX4090. 
Training is performed for 50 epochs using a batch size of 36, with the Adam optimizer and an initial learning rate of 0.0003. 
The primary loss function is the soft-label Kullback-Leibler divergence, implemented as a variant of cross-entropy for probabilistic (soft) targets. 

%
Regarding evaluation metric, the F1-score is computed every three iterations as obeying the challenge rule. To ensure the model is exposed to the entire training data distribution and to mitigate overfitting, the validation fold is rotated, keeping it separate from the training folds. The best-performing weights from each fold are then saved and used to test the provided challenge dataset.

\subsection{Experimental Results and Discussions}

To evaluate the performance of the backbone EfficientNetV2\_Large architecture, we conducted a comprehensive comparison against other deep-learning network architecture, including U-Net variants~\cite{le24}, conventional CNNs~\cite{incepv3,resnet,regnet,resnetmoco}, and transformer-based architectures~\cite{davit,vit,swin}. 
For a fair comparison, all models were initialized with pretrained weights from ImageNet dataset. 

\begin{figure}[h]
    \centering
    \includegraphics[width=0.8\linewidth]{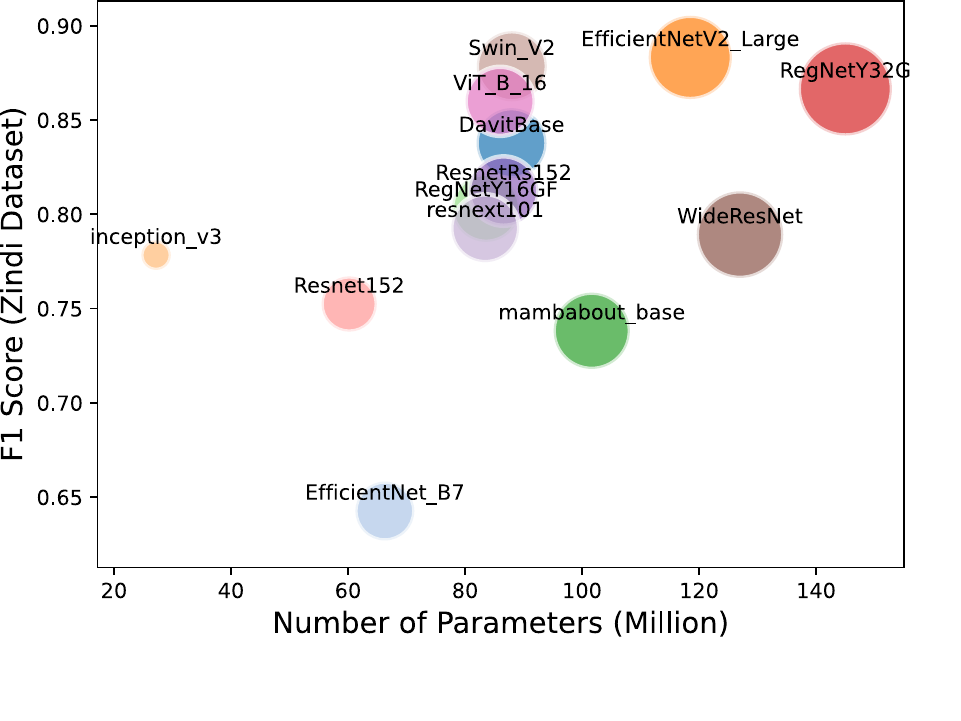}
    \vspace{-1.0 cm}
    \caption{Performance and parameter comparison among backbone deep neural network}
    \label{fig:model_performance}
\end{figure}

As illustrated in Fig.~\ref{fig:model_performance}, EfficientNetV2\_Large consistently outperforms other architectures on the test set, achieving the highest accuracy among architectures with comparable parameter sizes. 
Notably, although deeper models like ResNetY32G and WideResNet perform relatively well on training set, they exhibit a lower generalization gap compared to EfficientNetV2\_Large in test set. 
In contrast, InceptionV3, while shows efficient in other vision tasks, fails to generalize effectively under the current constraints. 
These results demonstrate that backbone deep-learning architecture plays a critical role in model generalization and robustness, even when other factors are held constant.

To analyze the impact of various hyperparameter configurations regarding EfficientNetV2\_Large network, we evaluate the number of input channels, normalization method, data augmentation techniques (SMOTE, MixUp, and CutMix), input dimension, and learning rate schedule. 
Table~\ref{tab:hyperparameter_performance} illustrates the result on test set of the Zindi dataset, where Baseline model is the standard EfficientNetV2\_Large without any modification. 
Each of following models was tested with one of additional features. 
The best model is the combination of all factors with the highest score of 0.8915.

\begin{table}[htbp]
\centering
\caption{Ablation Study on EfficientNetV2\_Large with training factors. Each symbol $^\ddagger$ indicates a modification from the baseline configuration.}
\label{tab:hyperparameter_performance}
\begin{tabular}{cccccccccc}
\toprule
\textbf{Model} & \textbf{Dim} & \textbf{\#Ch} & \textbf{Mixup} & \textbf{CutMix} & \textbf{LR Schedule} & \textbf{Norm.} & \textbf{SMOTE} & \textbf{Train F1} & \textbf{Test F1} \\
\midrule
Baseline        & 256 & 12 & No  & No  & No               & Standard & No  & 0.9398 & 0.8370 \\
EfficientNetV2  & 384$^\ddagger$ & 12 & Yes & No  & No               & Standard & No  & 0.9976 & 0.7866 (-0.05\textcolor{red}{\faArrowDown}) \\
EfficientNetV2  & 256 & 20$^\ddagger$ & Yes & Yes & No               & Standard & No  & 0.9469 & 0.8163 (-0.02\textcolor{red}{\faArrowDown})\\
EfficientNetV2  & 256 & 12 & Yes$^\ddagger$ & Yes$^\ddagger$ & No               & Standard & No  & 0.9891 & 0.8606 (+0.02\textcolor{blue}{\faArrowUp}) \\
EfficientNetV2  & 256 & 12 & Yes & Yes & Step$^\ddagger$             & Standard & No  & 0.9713 & 0.8379 (+0.0009\textcolor{blue}{\faArrowUp}) \\
EfficientNetV2  & 256 & 12 & Yes & Yes & CosineAnnealingLR$^\ddagger$ & Standard & No  & 0.9988 & 0.8729 (+0.04\textcolor{blue}{\faArrowUp}) \\
EfficientNetV2  & 256 & 12 & Yes & Yes & CosineAnnealingLR & Robust$^\ddagger$   & No  & 0.9976 & 0.8845 (+0.05\textcolor{blue}{\faArrowUp})\\
EfficientNetV2$^\star$  & 256 & 12 & Yes & Yes & CosineAnnealingLR & Standard & Yes$^\ddagger$ & 1.0000 & \textbf{0.8915} (+0.055\textcolor{blue}{\faArrowUp})\\
\bottomrule
\end{tabular}
\end{table}

To assess the effectiveness of the SVM classifier, we perform a series of comparative experiments using the best-performing model from Table~\ref{tab:hyperparameter_performance} as the baseline. 
This baseline model employs a traditional fully connected layer as its classification head. 
We then replace this layer with an SVM classifier and vary the regularization parameter $C$ to evaluate its impact on model generalization. 
Additionally, we evaluate the Light Gradient Boosting Machine (LGBM)\cite{lgbm}, a widely used machine learning algorithm, as an alternative classifier. The results, presented in Table \ref{tab:svm_comparison}, show that applying a small degree of regularization improves the model’s generalization performance.

\begin{table}[ht]
\centering
\caption{Comparison of EfficientNetV2 models using SVM as classifier with different configurations.}
\label{tab:svm_comparison}
\begin{tabular}{lccc}
\toprule
\textbf{Model}& \textbf{C} & \textbf{Train F1} & \textbf{Test F1} \\
\midrule
Baseline                    & - & 1.0000 & 0.8915 \\
EfficientNetV2+LGBM         & - & 1.0000 & 0.8826  \\
EfficientNetV2+SVM          & 1.00 & 1.0000 & 0.8840 \\
EfficientNetV2+SVM          & 0.75 & 1.0000 & 0.8894 \\
EfficientNetV2+SVM          & 0.5 & 1.0000 & 0.8921 \\
EfficientNetV2+SVM$^\star$  & 0.1 & 1.0000 & \textbf{0.8938} \\
\bottomrule
\end{tabular}
\end{table}

\begin{figure}[htbp]
    \centering
    \begin{subfigure}[b]{0.45\linewidth}
        \centering
        \includegraphics[width=\linewidth]{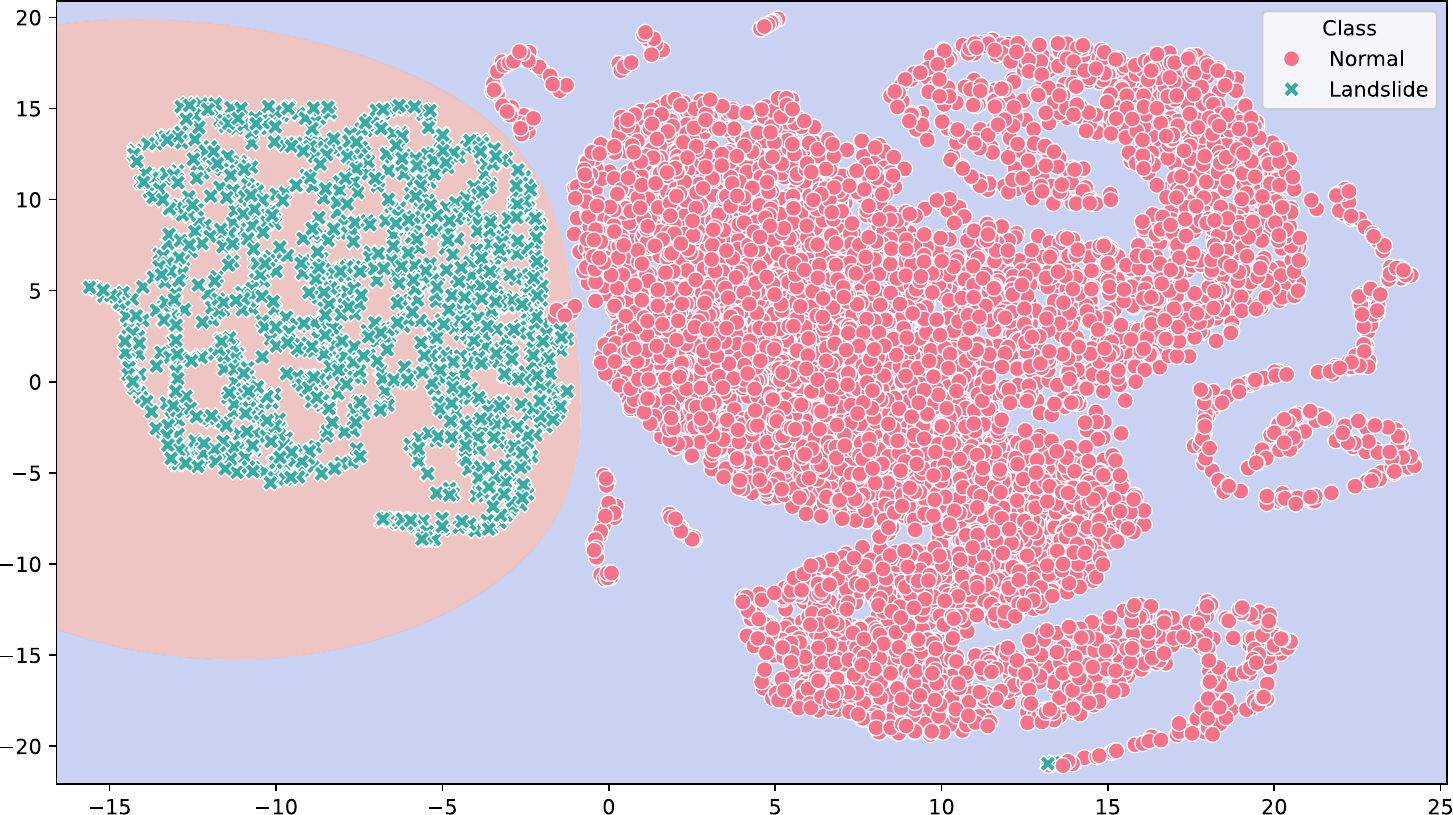}
        \caption{Feature visualization from the baseline model.}
        \label{fig:feat_base}
    \end{subfigure}
    \hfill
    \begin{subfigure}[b]{0.45\linewidth}
        \centering
        \includegraphics[width=\linewidth]{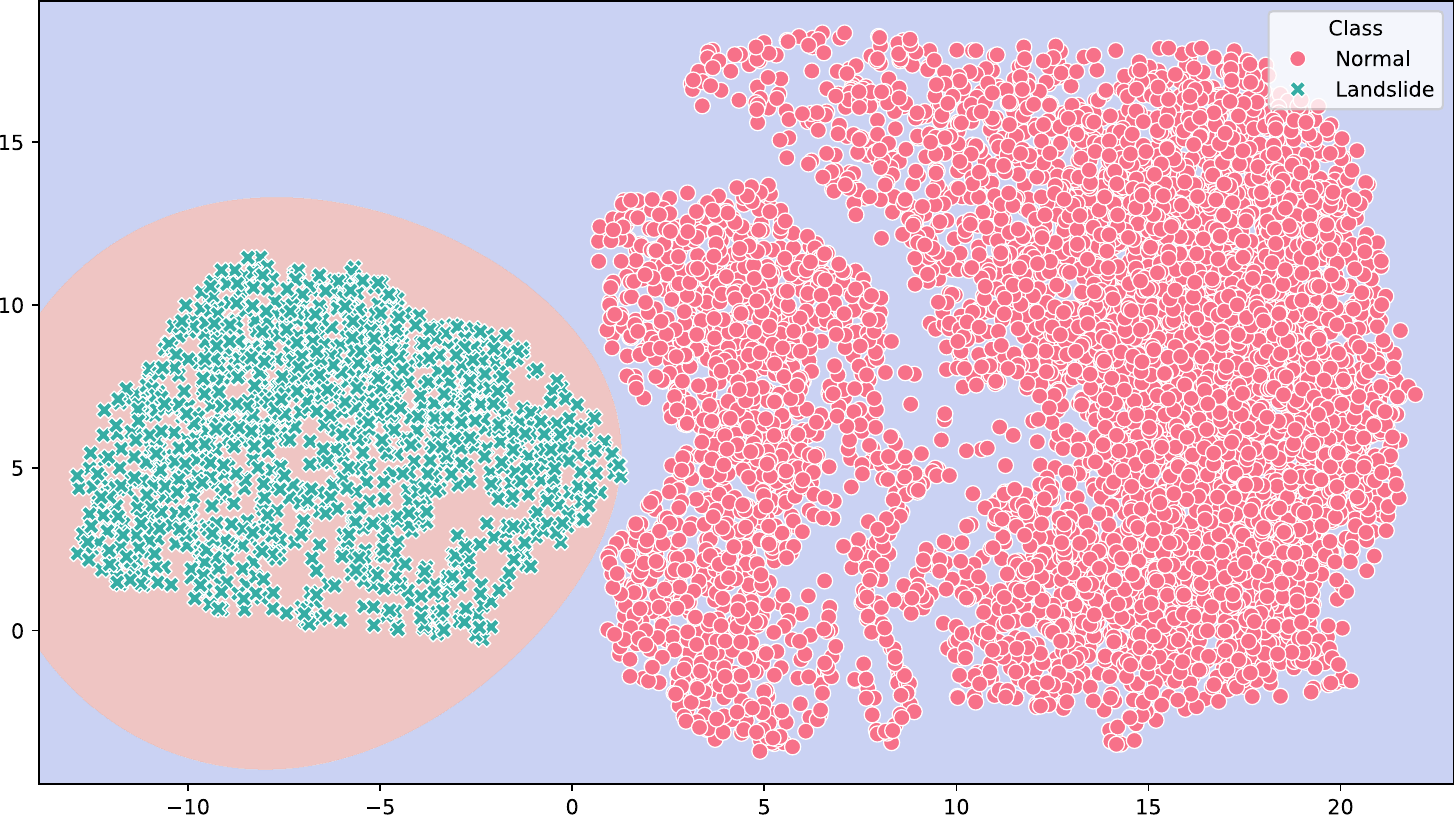}
        \caption{Feature visualization from the best model.}
        \label{fig:feat_svm}
    \end{subfigure}
    \caption{Comparison of embedding feature distributions using UMAP.}
    \label{fig:feature_aligment}
\end{figure}

Finally, to better understand the behavior of the models, we employ various explainable AI (xAI) tools. 
First, we apply Uniform manifold approximation and projection for dimension reduction(UMAP)~\cite{umap} to reduce the dimensionality of the embedding feature representations extracted from the network just before the classifier layer. 
Then, a simple SVM is used to illustrate the decision boundaries, as shown in Fig.~\ref{fig:feature_aligment}. 
The results demonstrate that using an SVM classifier as the final layer leads to more clearly separated feature regions compared to the traditional fully connected layer. This improved feature alignment may contribute to better generalization.

\begin{figure}[H]
    \centering
    \includegraphics[width=0.8\linewidth]{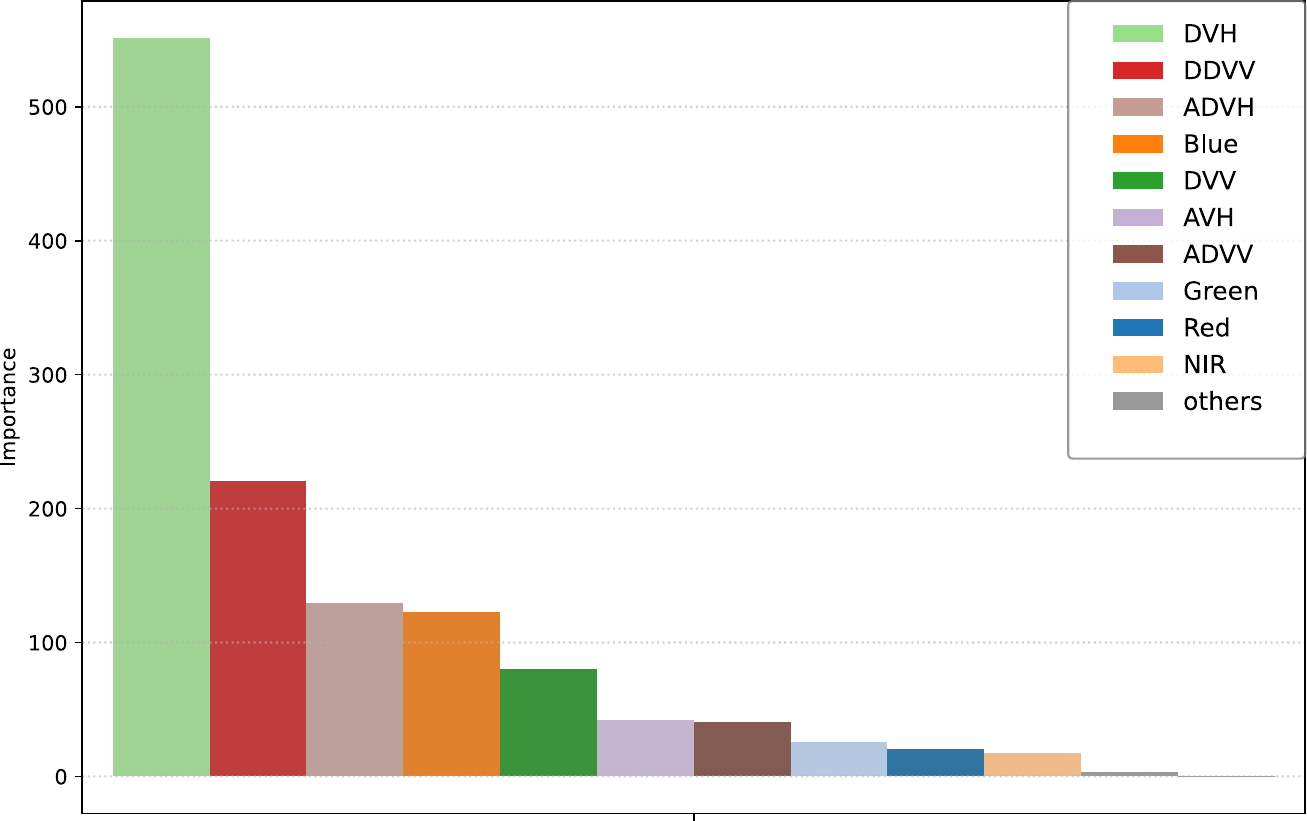}
    \caption{Feature importance obtained using the band-wise occlusion technique.}
    \label{fig:feature_importance}
\end{figure}

Second, we assess the importance of each input band among the original 12 bands by performing a manual occlusion analysis. Specifically, we systematically zero out each band by setting all pixel values in that band to zero—before passing the modified input through the trained model. The predicted probability for the landslide class is then compared to the original prediction (assumed to be 1.0 for landslide images). By accumulating the reduction in predicted probability for each occluded band, we estimate its relative contribution to the model’s predictive performance. The results of this analysis are illustrated in Fig.~\ref{fig:feature_importance}.

From this experiment, we observe that synthetic aperture radar (SAR) bands (e.g., DVH, DDVV, ADVH) contribute significantly more to the prediction than the optical bands (e.g., Red, Green, Blue, NIR). This finding is consistent with the fact that the dataset includes scenes under various conditions such as cloud cover or urban environments, which can degrade the effectiveness of optical bands in identifying landslide regions.

\section{Conclusion}
In this study, we proposed deep-learning framework that combines the EfficientNetV2\_Large architecture with a Support Vector Machine (SVM) classifier to improve landslide detection using multi-spectral satellite imagery. 
To address the class imbalance issue inherent in the dataset, we integrated a both offline and online data augmentation. 
Through extensive ablation studies, we demonstrated that incorporating augmentation techniques, robust deep-learning network, and SVM-based post processing yields superior performance, especially when fine-tuned with appropriate regularization to achieve an F1-score of 0.8938.
 Furthermore, explainability analyses using UMAP and band-wise occlusion confirmed that SAR-derived features play a more critical role than RGB/NIR bands in landslide identification, especially in complex environmental conditions.

%


\FloatBarrier

\bibliographystyle{splncs04}  
\bibliography{references}




\end{document}